\pdfoutput=1
\documentclass[11pt]{article}

\usepackage[]{EMNLP2022}

\usepackage{times}
\usepackage{latexsym}

\usepackage[T1]{fontenc}
\usepackage[utf8]{inputenc}

\usepackage{microtype}

\usepackage{inconsolata}
\usepackage{graphicx}
\usepackage{amsmath}
\usepackage{booktabs}
\usepackage{multirow}
\usepackage{amssymb}
\usepackage{algorithm}
\usepackage{algorithmic}

\usepackage{colortbl}
\newcolumntype{P}[1]{>{\centering\arraybackslash}p{#1}}

\usepackage[symbol]{footmisc}
\renewcommand{\thefootnote}{\fnsymbol{footnote}}

\usepackage{stfloats}
\usepackage{adjustbox}
\DeclareUnicodeCharacter{2212}{-}

\makeatletter
\g@addto@macro\normalsize{%
  \abovedisplayskip 1pt plus 1pt minus 1pt%
  \belowdisplayskip \abovedisplayskip
  \abovedisplayshortskip 2pt plus1pt  minus1pt%
  \belowdisplayshortskip 2pt plus1pt minus1pt%
}
\g@addto@macro\small{%
  \abovedisplayskip 2pt plus 1pt minus 1pt%
  \belowdisplayskip \abovedisplayskip
  \abovedisplayshortskip 2pt plus1pt  minus1pt%
  \belowdisplayshortskip 2pt plus1pt minus1pt%
}
\g@addto@macro\footnotesize{%
  \abovedisplayskip 2pt plus 1pt minus 1pt%
  \belowdisplayskip \abovedisplayskip
  \abovedisplayshortskip 2pt plus1pt  minus1pt%
  \belowdisplayshortskip 2pt plus1pt minus1pt%
}
\makeatother

\title{Self-Distillation with Meta Learning for Knowledge Graph Completion}
\author{Yunshui Li$^{1,2}$ \quad  Junhao Liu$^{1}$  \quad Chengming Li$^{3}$\footnotemark[1] \quad Min Yang$^{1}$\footnotemark[1] \\
        $^{1}$Shenzhen Institute of Advanced Technology, Chinese Academy of Sciences \\
        $^{2}$University of Chinese Academy of Sciences \\
        $^{3}$School of Intelligent Systems Engineering, Sun Yat-sen University\\
        \texttt{\{ys.li, jh.liu, min.yang\}@siat.ac.cn, lichengming@mail.sysu.edu.cn}
        }
\begin{document}
\maketitle
\renewcommand{\thefootnote}{\fnsymbol{footnote}}
\footnotetext[1]{Corresponding author}
\renewcommand{\thefootnote}{\arabic{footnote}}

\begin{abstract}
%Deep neural networks have achieved significant success in knowledge graph completion. In general, these deep models are typically based on high-dimensional embeddings, which require a large number of parameters to obtain optimal performance. Despite the effectiveness, high computational costs obstruct large deep models to be effectively deployed for real-time applications. In addition, the latest deep neural methods suffer from robust and generalization performance in the practical scenario, especially for the long-tail samples. 
In this paper, we propose a self-distillation framework with meta learning~(MetaSD) for knowledge graph completion with dynamic pruning, which aims to learn compressed graph embeddings and tackle the long-tail samples. Specifically, we first propose a dynamic pruning technique to obtain a small pruned model from a large source model, where the pruning mask of the pruned model could be updated adaptively per epoch after the model weights are updated. The pruned model is supposed to be more sensitive to difficult-to-memorize samples (e.g., long-tail samples) than the source model. 
Then, we propose a one-step meta self-distillation method for distilling comprehensive knowledge from the source model to the pruned model, where the two models co-evolve in a dynamic manner during training.
%In particular, we alternately update the pruned model based on the output of the source model and optimize the source model based on the pruned model's performance via meta learning. 
\fnsymbol{footnote}
In particular, we exploit the performance of the pruned model, which is trained alongside the source model in one iteration, to improve the source model's knowledge transfer ability for the next iteration via meta learning.
Extensive experiments show that MetaSD achieves competitive performance compared to strong baselines, while being 10x smaller than baselines. \footnote{The dataset and code are publicly available at \url{https://github.com/pldlgb/MetaSD}}
\end{abstract}

\section{Introduction}
Knowledge graphs (KG) have become the key technology to represent structural relations between entities and play an important role in question answering \cite{shen2018knowledge}, dialogue systems \cite{yan2017building}, and entity disambiguation \cite{mulang2020evaluating,si2022scl,si2023santa}.
However, most KGs are growing at a rapid pace and are far from complete. Therefore, it is necessary to develop knowledge graph completion (KGC) approaches to add missing triples to the KGs, so as to improve the quality of KGs. 

Recent advances in KGC primarily work on knowledge graph embedding (KGE) by converting the entities and relations in KGs into low-dimensional vectors. %that capture the rich semantic information of the entities and relations. 
Early studies on KGE introduce a margin-based pairwise ranking function to measure the Euclidean distance or similarity between the relational projection of entities \cite{nickel2011three,bordes2013translating,yang2014embedding,trouillon2016complex}. Among them, TransE \cite{bordes2013translating} is the most widely used KGE method, which views the relation as translation from a head entity to a tail entity. 
%However, these models only explore structure information of observed triple facts, which suffer from the sparseness of KGs.
Recently, neural networks, such as neural tensor network (NTN) \cite{socher2013reasoning} and neural association model (NAM) \cite{liu2016probabilistic} have been proposed to encode semantic matching and achieved remarkable predictive performance for KGC. 

\begin{figure}[t]
\centering % 图片居中
\includegraphics[width = 0.35\textwidth]{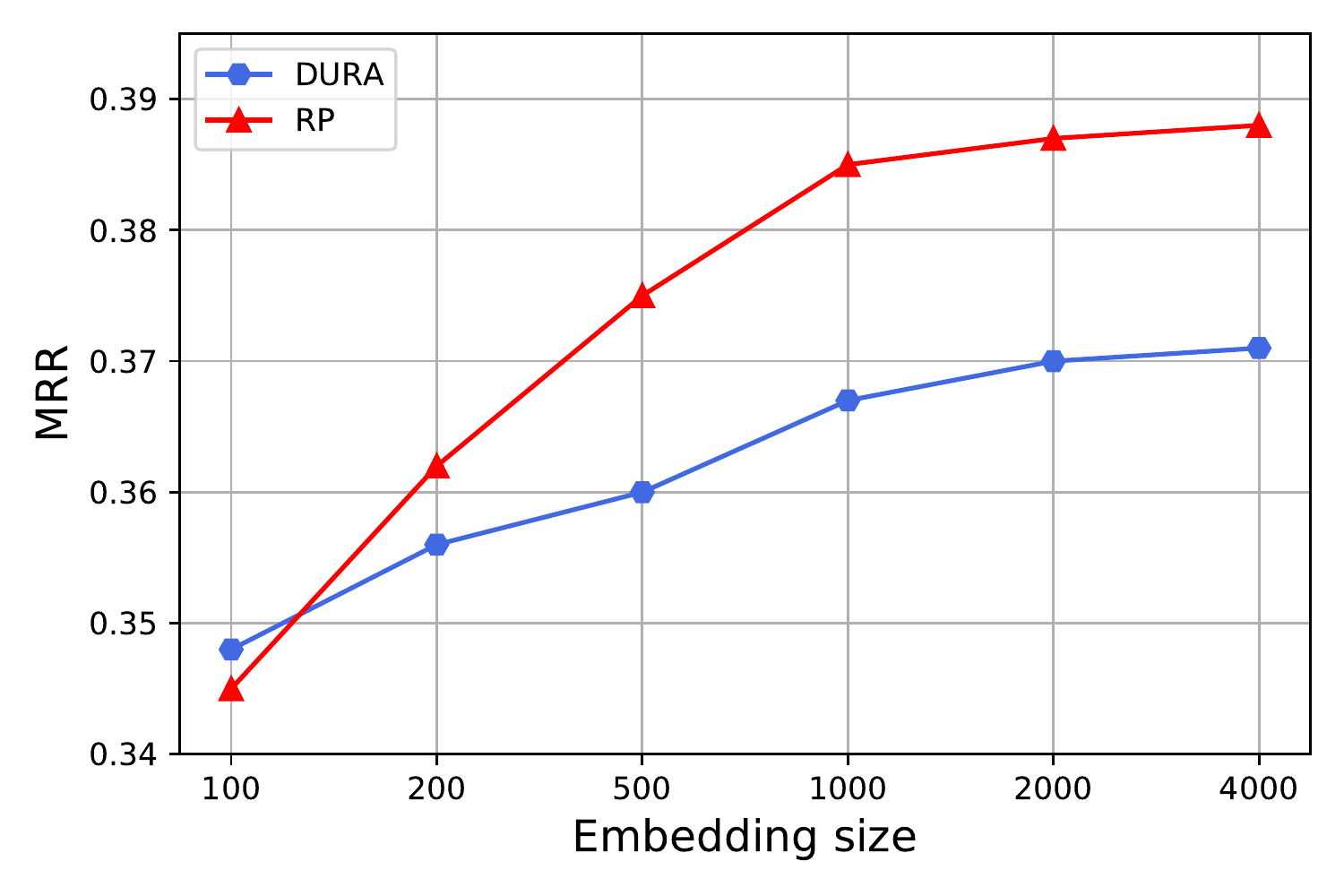}
\caption{The MRR scores w.r.t. the graph embedding sizes of DURA and RP on FB15k-237.}
\label{fig:fbdim}
\end{figure}

To increase the capacity of the KGE models, a larger embedding size with more parameters is a common technique in practice. As shown in Figure~\ref{fig:fbdim}, the prediction performance of the KGC models such as DURA~\citep{zhang2020duality} and RP~\citep{chen2021relation} can be largely improved by increasing the graph embedding size. Although the large graph embedding often bring obvious performance improvements, it may also become the major obstacle for model deployment and real-time prediction, especially for memory-limited and resource-constrained devices.

In addition, the distribution of samples with long-tail is prevalent in KGs \cite{zhang2019long}, where a large portion of relations have much fewer triples than other relations. However, most previous studies mainly focus on the predictive performance on overall test data, without taking long-tail samples into consideration. These models suffer from robust and generalization performance in the practical scenario. 
%In particular, the latest neural models suffer from robust and generalization performance in the practical scenario, especially for the long-tail samples. 
Although several recent works \cite{xiong2018one,sheng2020adaptive} have been proposed for few-shot KGC, these models are not adapted to the model compression frameworks.

In this paper, we propose a self-distillation framework with meta learning~(MetaSD) for knowledge graph completion with dynamic pruning. First, we propose a dynamic pruning technique to obtain a small pruned model from a large source model at the start of each training epoch. Concretely, the pruning mask of the pruned model could be updated adaptively per epoch after updating the model weights. The pruned model is supposed to be more sensitive to the difficult-to-memorize samples (e.g., long-tail samples) than the source model. 
Second, we propose a one-step meta self-distillation method to distill comprehensive knowledge from the source model to the pruned model, where the two models co-evolve in a dynamic manner during the whole knowledge distillation process.
The key idea is to use the performance of the pruned model, which is trained alongside the source model in one iteration, to improve the source model for the next iteration by borrowing the idea of learning to learn from meta learning \cite{finn2017model}. 
%In particular, we alternately update the pruned model based on the output of the source model and optimize the source model based on the pruned model's performance via meta learning. 
%In particular, we exploit the performance of the pruned model, which is trained alongside the source model in one iteration, to improve the source model's knowledge transfer ability for the next iteration by borrowing the idea of learning to learn from meta learning.
In particular, we define the objectives of the source model as functions of the pruned model’s performance on a quiz set. The usage of ``gradient by gradient'' strategy makes the source model adjust to the learning state of the pruned model, and improves both the source and pruned models.  

The main contributions of our method can be three-fold. 
(1) We propose a self-distillation framework to compress KG embeddings for KGC. The source and pruned models co-evolve in a dynamic manner during training, thus we can avoid pre-training a large model in advance, and the performance of the pruned model is not limited to that of a pre-trained large model.
(2) We exploit the feedback from the pruned model to guide the source model with meta learning, making the source model transfer better knowledge to the pruned network. 
(3)  Experimental results on two benchmark datasets show that our model achieves competitive performance compared to strong baselines, while being 10x smaller than other KGC models. %In addition, our method also achieves better prediction results than the compared knowledge distillation methods.  

\begin{figure*}[]
\centering % 图片居中
\includegraphics[width = 16cm]{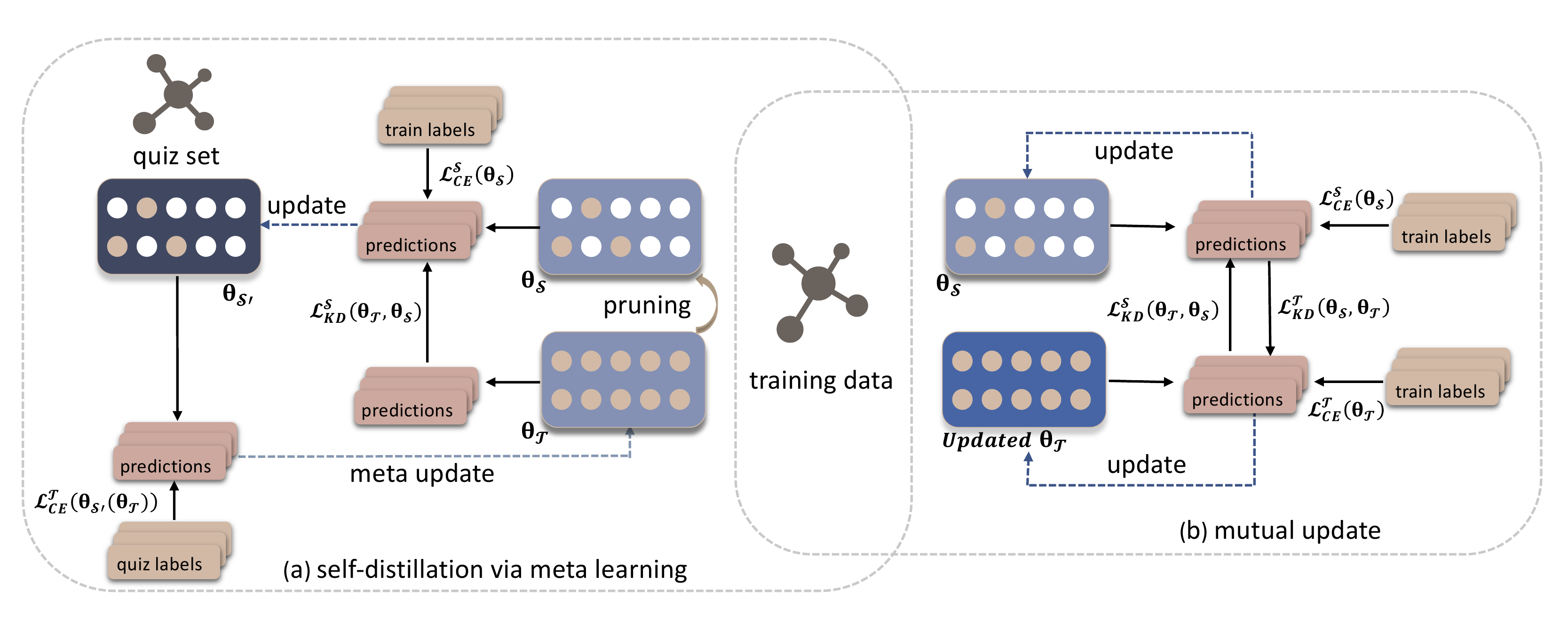}
\caption{The overview of MetaSD framework. (a) We prune the teacher $\mathcal{T}$ to obtain the student $\mathcal{S}$ and perform knowledge distillation on training data to update the temporary copy $\mathcal{S}'$ from $\mathcal{S}$. Then, the source model $\mathcal{T}$ is optimized based on the feedback of $\mathcal{S}'$ on a held-out quiz set $\mathcal{Q}$;$\ $ (b) We discard $\mathcal{S}'$ and optimize  the meta-updated $\mathcal{T}$ and real $\mathcal{S}$ alternately by performing mutual learning on the training data.}
\label{fig:figure1}
\end{figure*}

\section{Methodology}
\paragraph{Problem Definition}
Suppose a KG can be viewed as a graph $\mathcal{G}=\{(h, r, t)\}\in E\times R\times E$, where $E$ and $R$ represent the entity (node) set and relation (edge) set respectively. $(h, r, t)$ represents a triple, where $h$, $t$ and $r$ indicate head entity, tail entity and the relation between two entities, respectively. Given the KG $\mathcal{G}$, the goal of KGC is to infer missing links
based on existing triples in the KG.

\paragraph{Model Overview}
The overview of our MetaSD method is illustrated in Figure~\ref{fig:figure1}. 
We adopt ComplEx~\citep{trouillon2016complex} as our backbone model, which is treated as the source model $\mathcal{T}$. First, we use a magnitude-based weight pruning method~\citep{han2015deep,zhu2017prune} to obtain a pruned model $\mathcal{S}$ from the source model $\mathcal{T}$. Second, we propose a one-step meta self-distillation method for distilling comprehensive knowledge from the source model $\mathcal{T}$ to the pruned model $\mathcal{S}$, where the two models co-evolve in a dynamic manner during training. Next, we introduce the proposed MetaSD method in detail. 

\subsection{Network Pruning}
We use the magnitude-based weight pruning method~\citep{han2015deep,zhu2017prune} to create a self-competitive compressed model $\mathcal{S}$ by pruning the source model $\mathcal{T}$. In particular, 
we fix the pruning rate $\gamma$ during the whole training process. At each iteration, we first calculate the sum of parameter numbers of all layers be pruned as $P$, and sort all the weights by their absolute values. Then, we prune a certain fraction (i.e., $\gamma$) of weights that have lowest absolute weight values. In particular, to dynamically adjust the pruned network $\mathcal{S}$ during each iteration, we prune the chosen weight by setting the corresponding values in a binary mask to zero, instead of directly setting the weights to zero. 

\subsection{Self-Distillation via Meta Learning}
We exploit the performance of the pruned model, which is trained alongside the source model in one iteration, to improve the source model's knowledge transfer ability for the next iteration via meta learning. 
In particular, we alternately update the pruned model $\mathcal{S}$ based on the output of the source model $\mathcal{T}$ and optimize the source model $\mathcal{T}$ based on the pruned model's performance via meta learning. 

\paragraph{Model $\mathcal{S}$ with Knowledge Distillation}
Formally, we use the function $\mathcal{S}(x_i; \theta_{\mathcal{S}})$ to denote the soft prediction of the compressed model, where $\theta_{\mathcal{S}}$ represents the parameters of the pruned model $\mathcal{S}$. We calculate the cross-entropy loss $\mathcal{L}^{\mathcal{S}}_{\rm CE}(\theta_{\mathcal{S}})$ on the training data in current batch as:
\begin{equation}
\mathcal{L}^{\mathcal{S}}_{\rm CE}(\theta_{\mathcal{S}})= \frac{1}{N}\sum_{i=1}^{N}{\rm CE}(y_i, \mathcal{S}(x_i; \theta_{\mathcal{S}}))
\end{equation}
where $N$ denotes the number of training samples. ${\rm CE(\cdot)}$ represents the cross-entropy function. 

To further improve the performance of $\mathcal{S}$, we also design a knowledge distillation loss  $\mathcal{L}^{\mathcal{S}}_{\rm KD}$ that encourages the output of $\mathcal{S}$ to mimic that of $\mathcal{T}$. In particular, we minimize the Kullback-Leibler Divergence (KL-divergence) between the output distributions of $\mathcal{S}$ and $\mathcal{T}$ by:
\begin{equation}
\resizebox{0.89 \hsize}{!}{%
$\mathcal{L}^{\mathcal{S}}_{\rm KD}(\theta_{\mathcal{S}},\theta_{\mathcal{T}})= \frac{1}{N}\sum_{i=1}^{N}{\rm KL}\big(\mathcal{S}(x_i; \theta_{\mathcal{S}})||\mathcal{T}(x_i; \theta_{\mathcal{T}})\big) $}
\end{equation}
where $\theta_{\mathcal{T}}$ represents the parameters of  $\mathcal{T}$.

The cross-entropy loss $\mathcal{L}^{\mathcal{S}}_{\rm CE}$ and the knowledge distillation loss $\mathcal{L}^{\mathcal{S}}_{\rm KD}$ are combined to form the overall loss $\mathcal{L}_{\mathcal{S}}$ for the compressed model $\mathcal{S}$ as:
\begin{equation}
\resizebox{0.89 \hsize}{!}{%
$\mathcal{L}_{\mathcal{S}}(\theta_{\mathcal{S}},\theta_{\mathcal{T}})= \alpha \mathcal{L}^{\mathcal{S}}_{\rm CE}(\theta_{\mathcal{S}}) + (1-\alpha) \mathcal{L}^{\mathcal{S}}_{\rm KD}(\theta_{\mathcal{S}},\theta_{\mathcal{T}}) $}
\end{equation}
where $\alpha$ is a hyperparameter to balance the relative importance of the two loss functions.

\paragraph{Model $\mathcal{T}$ with Meta Learning}
We exploit feedback from the compressed model's learning
state to improve the source model's knowledge transfer ability throughout the distillation process, instead of keeping the source model $\mathcal{T}$ fixed in the training process. 
We train both $\mathcal{T}$ and $\mathcal{S}$ in an iterative manner until convergence. This interaction between the two models can be seen as a form of meta learning with a bi-level optimization process, which comprises three steps: \textbf{Virtual-Train}, \textbf{Meta-Train}, and \textbf{Actual-Train} \cite{xu2021faster}.
%Inspired by meta learning, we update the objective of the source model $\mathcal{T}$ through reusing the parameter update of the compressed model $\mathcal{S}$ learned.  
That is, the compressed model $\mathcal{S}$ is the inner-learner and the source model $\mathcal{T}$ is the meta-learner. 

For each training step,
we first copy the parameters $\theta_{\mathcal{S}}$ of the compressed model $\mathcal{S}$ to a ``virtual'' compressed model $\mathcal{S}'$, and then update the parameters $\theta_{\mathcal{S}}^{'}(\theta_{\mathcal{T}})$ of the ``virtual'' compressed model $\mathcal{S}'$ with SGD \cite{bottou2012stochastic} for the \textbf{Virtual-Train} as:
%Here, we use $\theta_{\mathcal{S}}^{'}(\theta_{\mathcal{T}})$ to present the optimal parameters of $\mathcal{S}$ after being trained on the last iteration:
\begin{equation}
\label{equation:S-update-2}
  \theta_{\mathcal{S}}^{'}(\theta_{\mathcal{T}}) = \theta_{\mathcal{S}} - \lambda \nabla_{\theta_{\mathcal{S}}} \mathcal{L}_{\mathcal{S}}(\theta_{\mathcal{S}},\theta_{\mathcal{T}})
\end{equation}

Then, the source model $\mathcal{T}$ is optimized based on the feedback of $\mathcal{S}$ on a held-out quiz set $\mathcal{Q}$. We perform a derivative over a derivative (a Hessian matrix) to update $\theta_{\mathcal{T}}$, by using a retained computational graph of $\theta_{\mathcal{S}}^{'}$ 
in order to compute derivatives with respect
to $\theta_{\mathcal{T}}$. The source model $\mathcal{T}$ is optimized by minimizing the cross-entropy loss over the quiz set $Q$  for the \textbf{Meta-Train} as:
\begin{equation}
\label{equation:theta-T-update}
\resizebox{0.89 \hsize}{!}{%
$\mathcal{L}^{\mathcal{T}}_{\rm CE}\big(\theta_{\mathcal{S}}^{'}\big(\theta_{\mathcal{T}})\big) = \frac{1}{M}\sum_{i=1}^{M}{\rm CE}(y'_i, \mathcal{S}(x'_i; \theta_{\mathcal{S}})\big)$}
\end{equation}
where $M$ is the training samples in the quiz set $Q$. $x'$ and $y'$ denote the input sample and corresponding label in quiz set $q\in \mathcal{Q}$, respectively. 

Finally, we update the source model $\mathcal{T}$ with SGD \cite{bottou2012stochastic} as follows:
\begin{equation}
\label{equation:theta-T-update}
  \theta_{\mathcal{T}} \gets \theta_{\mathcal{T}}-\mu \nabla_{\theta_{\mathcal{T}}} \mathcal{L}^{\mathcal{T}}_{\rm CE}(\theta_{\mathcal{S}}^{'}\big(\theta_{\mathcal{T}})\big)
\end{equation}
where $\mu$ is the learning rate for the Meta-Train. 

\paragraph{Mutual Update of $\mathcal{T}$ and $\mathcal{S}$ for Self-Distillation} 
In our self-distillation framework, the source model $\mathcal{T}$ and the compressed model $\mathcal{S}$ co-evolve in a dynamic manner during the whole KD process. Instead of updating $\mathcal{T}$ with cross-entropy loss, we learn both $\mathcal{T}$ and $\mathcal{S}$ models mutually. % in an iterative manner until convergence. 

Formally, for the \textbf{Actual-Train}, we first update the compressed model's parameters $\theta_{\mathcal{S}}$ with the training data and the updated parameters $\theta_{\mathcal{T}}$ as:
\begin{equation}
\theta_{\mathcal{S}} = \theta_{\mathcal{S}} - \lambda \nabla_{\theta_{\mathcal{S}}} \mathcal{L}_{\mathcal{S}}(\theta_{\mathcal{S}},\theta_{\mathcal{T}})
\label{eq:sgd-S}
\end{equation}

The source model $\mathcal{T}$ is also optimized by the combination of the cross-entropy loss $\mathcal{L}^{\mathcal{T}}_{\rm CE}$ and the knowledge distillation loss $\mathcal{L}^{\mathcal{T}}_{\rm KD}$ as:
\begin{equation}
\label{equation:theta-T-update}
\resizebox{0.8 \hsize}{!}{%
$ \mathcal{L}^{\mathcal{T}}_{\rm CE}(\theta_{\mathcal{T}}) = \frac{1}{N}\sum_{i=1}^{N}{\rm CE}(y_i, \mathcal{T}(x_i; \theta_{\mathcal{T}})\big) $}
\end{equation}
\vspace{-0.4cm}
\begin{equation}
\resizebox{0.89 \hsize}{!}{%
$\mathcal{L}^{\mathcal{T}}_{\rm KD}(\theta_{\mathcal{S}},\theta_{\mathcal{T}})= \frac{1}{N}\sum_{i=1}^{N}{\rm KL}\big(\mathcal{S}(x_i; \theta_{\mathcal{S}})||\mathcal{T}(x_i; \theta_{\mathcal{T}})\big) $}
\end{equation}
\vspace{-0.4cm}
\begin{equation}
\resizebox{0.84 \hsize}{!}{%
$\mathcal{L}_{\mathcal{T}}(\theta_{\mathcal{S}},\theta_{\mathcal{T}})= \beta \mathcal{L}^{\mathcal{T}}_{\rm CE}(\theta_{\mathcal{S}}) + (1-\beta) \mathcal{L}^{\mathcal{T}}_{\rm KD}(\theta_{\mathcal{S}},\theta_{\mathcal{T}}) $}
\end{equation}
where $\beta$ is a hyperparameter to balance the relative importance of the two loss functions. We first update the source model's parameters $\theta_{\mathcal{T}}$ with the training data and the updated parameters $\theta_{\mathcal{S}}$ as:
\begin{equation}
\theta_{\mathcal{T}} = \theta_{\mathcal{T}} - \lambda \nabla_{\theta_{\mathcal{T}}} \mathcal{L}_{\mathcal{T}}(\theta_{\mathcal{S}},\theta_{\mathcal{T}})
\label{eq:sgd-T}
\end{equation}

We train the source and compressed models in an iterative manner until convergence. 
Overall, the self-distillation with meta learning is defined in Algorithm~\ref{alg1}.

\begin{algorithm}[H]
% 	\textsl{}\setstretch{1.8}
% 	\renewcommand{\algorithmicrequire}{\textbf{Input:}}
	\renewcommand{\algorithmicensure}{\textbf{Output:}}
	\caption{Self-Distillation with Meta Learning}
	\label{alg1}
	\begin{algorithmic}[1]
		\REQUIRE train set $\mathcal{D}$, quiz set $\mathcal{Q}$, source model $\theta_{\mathcal{T}}$
		\REQUIRE learning rate $\lambda$, learning rate $\mu$, $i\leftarrow 0$
		\REPEAT
		%\For{each training iteration }{
		\STATE $i\leftarrow i+1$
		\STATE Sample a batch of training data $x$ from $\mathcal{D}$
		\STATE Get pruned model $\theta_{\mathcal{S}}$ by pruning $\theta_{\mathcal{T}}$ 
		\STATE Copy pruned model's parameters $\theta_{\mathcal{S}}$ to a\\
		~~``virtual'' pruned model $\theta_{\mathcal{S}}^{'}$
		%Pruning source model $\theta_T$ and get pruned model $\theta_S$ and copy model $\theta_{S'}$: \\
		  % \hspace{\algorithmicindent} $\theta_S = MagnitudePruning(\theta_T)$\\
		  % \hspace{\algorithmicindent}$\theta_{S'} = Copy(\theta_S)$
		\STATE Update $\theta_{\mathcal{S}}^{'}$ with $x$ and $\theta_{\mathcal{T}}$: \# \textit{Virtual-Train} \\ \hspace{\algorithmicindent} $\theta'_{\mathcal{S}}(\theta_{\mathcal{T}}) = \theta_{\mathcal{S}} - \lambda \nabla_{\theta_{\mathcal{S}}} \mathcal{L}_{\mathcal{S}}(x;\theta_{\mathcal{S}},\theta_{\mathcal{T}})$
		\STATE Sample a batch of quiz data $q$ from $\mathcal{Q}$
		\STATE Update $\theta_{\mathcal{T}}$ with $q$ and $\theta'_{\mathcal{S}}$: \# \textit{Meta-Train} \\ \hspace{\algorithmicindent} $\theta_{\mathcal{T}} \leftarrow \theta_{\mathcal{T}} - \mu \nabla_{\theta_{\mathcal{T}}} \mathcal{L}^{\mathcal{T}}_{\rm CE}(q;\theta'_{\mathcal{S}}(\theta_{\mathcal{T}}))$
		\STATE Mutual update $\theta_{\mathcal{T}}$ and $\theta_{\mathcal{S}}$: \# \textit{Actual-Train} \\
		    \hspace{\algorithmicindent}$\theta_{\mathcal{S}}(\theta_{\mathcal{T}}) = \theta_{\mathcal{S}} - \lambda \nabla_{\theta_{\mathcal{S}}} \mathcal{L}_{\mathcal{S}}(x;\theta_{\mathcal{S}};\theta_{\mathcal{T}})$\\
		    \hspace{\algorithmicindent} $\theta_{\mathcal{T}}(\theta_{\mathcal{S}}) = \theta_{\mathcal{T}} - \lambda \nabla_{\theta_{\mathcal{T}}} \mathcal{L}_{\mathcal{T}}(x;\theta_{\mathcal{S}};\theta_{\mathcal{T}})$
		\UNTIL $i$ == \textit{max iterations}
		%}
		\ENSURE source model $\theta_{\mathcal{T}}$ and pruned model $\theta_{\mathcal{S}}$
	\end{algorithmic}  
\end{algorithm}

\begin{table*}
    \centering
    \resizebox{1.0\textwidth}{!}{
    \begin{tabular}{p{2.0cm}P{1.4cm}P{1.4cm}P{1.4cm}P{1.4cm}P{1.4cm}P{1.4cm}P{1.4cm}P{1.4cm}P{1.4cm}P{1.4cm}P{1.4cm}}
    % \toprule
    \toprule
    \multirow{2}{*}{\textbf{Model}}& \multicolumn{5}{c}{\textbf{FB15k-237}}&
    \multicolumn{5}{c}{\textbf{WN18RR}} &\multirow{2}{*}{\textbf{Dim}}\\
    \cmidrule(lr){2-6}
    \cmidrule(lr){7-11}
    & MRR & Hits@1 & Hits@3 & Hits@10 & \textbf{Size} & MRR & Hits@1 & Hits@3 & Hits@10 & \textbf{Size} & \\
    \hline
    \hline
    TransE $\star$ & 0.313&0.221&0.347&0.497&- & 0.228& 0.053& 0.368&0.520 &-&-\\ %\citep{bordes2013translating}
    RotatE $\star$ & 0.333&0.240&0.368&0.522 &- & 0.478& 0.439& 0.494& 0.553 &-&- \\%\citep{sun2019rotate}
    % \hline
    CP & 0.333 & 0.247& 0.360& 0.508 &50M &0.438 &0.414 &0.444& 0.485 &156M&2k\\
    RESCAL  & 0.353&0.264&0.383&0.528 &125M & 0.455& 0.419&0.460 &0.493 &26M&-\\%\cite{nickel2011three}
    ComplEx  &0.346&0.256 & 0.370 &0.525 &60M &0.460 &0.428 &0.475 &0.522 &156M&2k \\%\citep{trouillon2016complex}
    DURA  & 0.371 & 0.276 & 0.408 & 0.560 & 60M  & \textbf{0.491} & \textbf{0.449}  &0.503&\textbf{0.571}&156M&2k\\%\citep{zhang2020duality}
    RP &0.388&0.298&0.425&0.568&60M &0.488&0.443&\textbf{0.505}&0.568&156M&2k \\%\citep{chen2021relation}
    % % \hline
    KD &0.371&0.282&0.408&0.550&6M &0.470&0.427&0.485&0.530&15M&0.2k\\
    DML &0.373&0.280&0.410&0.563&6M &0.472&0.429&0.485&0.535&15M&0.2k\\
    \rowcolor{gray!20}MetaSD & \textbf{0.391}&\textbf{0.300}&\textbf{0.428}&\textbf{0.571} &6M&\textbf{0.491}&\underline{0.447}&\underline{0.504}&\underline{0.570} &15M&0.2k\\
    % \rowcolor{gray!20}PTSD & \textbf{0.395}&\textbf{0.304}&\textbf{0.432}&\textbf{0.574}&60M&\textbf{0.492}&\textbf{0.450}&\textbf{0.505}&\textbf{0.571}&156M&2k\\
    \bottomrule
    \end{tabular} }
    \caption{\label{benchmark}
    Experimental results on FB15k-237 and WN18RR test sets for KGC. The results with $\star$ are taken from LibKGE~\citep{libkge}. CP, ComplEx and RESCAL are implemented by following~\citep{zhang2020duality}. %and we achieve better results than that in the original paper. 
    }
\end{table*}
% #######################################################################

\section{Experimental Setup}
\subsection{Datasets}
We conduct experiments on two KGC benchmark datasets: WN18RR~\citep{toutanova2015observed} and FB15k-237~\citep{dettmers2018convolutional}. WN18RR consists of 40,943 entities and 11 relations, and there are 86k/3k/3k instances for training/validation/testing respectively. FB15k-237 contains 14,541 entities and 237 relations, and there are 272k/17k/20k instances for training/validation/testing.

%WN18RR and FB15k-237 are subsets of WN18~\citep{bordes2013translating} and FB15k~\citep{bordes2013translating}, created by removing the inverse relations from validation and test sets. Table~\ref{tab:dataset} shows the statistics of these two datasets.

\subsection{Baseline Methods}
We compare  MetaSD with several strong KGC baselines, including CP~\citep{hit1927},  RESCAL~\citep{nickel2011three}, TransE~\citep{bordes2013translating}, ComplEx~\citep{trouillon2016complex},
RotatE~\citep{sun2019rotate},
DURA~\citep{zhang2020duality}, and RP~\citep{chen2021relation}.  We also compare MetaSD with two widely used KD methods: knowledge distillation (KD)~\citep{hinton2015distilling} and deep mutual learning (DML)~\citep{zhang2018deep}, where the pre-trained RP~\citep{chen2021relation} is used as their teacher model. 

\subsection{Implementation Details}
The source model of MetaSD is initialized with ComplEx~\citep{trouillon2016complex}, following the previous work \citep{zhang2020duality}. Similar to~\citet{chen2021relation}, we add relation prediction as an auxiliary task. We set the pruning rate $\gamma$ to 0.9 to strike a balance between the effectiveness and efficiency of the model.
We set balance hyperparameters $\alpha = \beta = 0.5$. We choose Adagrad~\citep{duchi2011adaptive} as the optimizer and the learning rate $\mu$ to $1e^{-4}$ and $\lambda$ to $1e^{-1}$. The quiz set is randomly sampled from training data and then fixed. We adopt widely used \textit{filtered} evaluation metrics of mean reciprocal rank~(MRR), Hits@1, Hits@3, and Hits@10 as described in~\cite{bordes2013translating}.

\section{Experimental Results}
\subsection{Overall Performance}
As shown in Table~\ref{benchmark}, we report the results of the compressed model (denoted as MetaSD). Note that the parameters of RASCAL are proportional to the square of the number of relations, resulting in large differences in size between the two datasets. We observe that MetaSD achieves competitive performance compared to other high-dimensional baseline models on the two datasets for KGC, while being 10x smaller than baseline methods. 
In addition, MetaSD also outperforms than two widely used KD methods that have the same size and dimension with MetaSD.

\subsection{Long-tail Evaluation}
We investigate the effectiveness of MetaSD for dealing with the long-tail samples. In particular, we collect the long-tail samples from the FB15k-237 test set by choosing the relations that have fewer than 1000 training instances. In total, there are 187 relations, which accounts for 79\% of the total relation types but only 24\% of train set.  Table~\ref{longtail} reports the results of MetaSD and compared methods on the long-tail set. MetaSD significantly outperforms other models on the long-tail samples, which verifies the effectiveness of MetaSD in tackling the long-tail samples. 
\begin{table}
\centering
\resizebox{0.45\textwidth}{!}{
\begin{tabular}{p{1.8cm}P{1.2cm}P{1.2cm}P{1.2cm}P{1.2cm}}
    \toprule
    % \hline
    \textbf{Model} & \textbf{MRR} & \textbf{H@1} & \textbf{H@3} & \textbf{H@10} \\
    \hline\hline
    DURA & 0.452 & 0.354 & 0.498 & 0.644\\
    RP & 0.462& 0.372& 0.504& 0.645 \\
    % \hline
    \rowcolor{gray!20}MetaSD-$\mathcal{T}$ & 0.468 & 0.380 & 0.510 & 0.642\\
    \rowcolor{gray!20}MetaSD & \textbf{0.471} & \textbf{0.381} & \textbf{0.512} & \textbf{0.646} \\
    \bottomrule
\end{tabular}
}
\caption{\label{longtail}
Results on long-tail data from FB15k-237.
}
\end{table}
\begin{table}
\centering
\resizebox{0.45\textwidth}{!}{
\begin{tabular}{p{1.8cm}P{1.2cm}P{1.2cm}P{1.2cm}P{1.2cm}}
    \toprule
    % \hline
    \textbf{Model} & \textbf{MRR} & \textbf{H@1} & \textbf{H@3} & \textbf{H@10} \\
    \hline\hline
    \textbf{MetaSD} & \textbf{0.391} & \textbf{0.300} & \textbf{0.428} & \textbf{0.571}\\
    \ \ \  w/o P & 0.381& 0.292& 0.415& 0.561 \\
    \ \ \ w/o M & 0.378& 0.287&0.412& 0.555\\
    \ \ \ w/o P\&M & 0.373 & 0.280 & 0.410 &0.563\\
    % \hline
    % \rowcolor{gray!20}MetaSD-$\mathcal{T}$ & 0.468 & 0.380 & 0.510 & 0.642\\
    % \rowcolor{gray!20}MetaSD & \textbf{0.471} & \textbf{0.381} & \textbf{0.512} & \textbf{0.646} \\
    \bottomrule
\end{tabular}
}
\caption{\label{ablation}
Results of ablation study on FB15k-237. $P$ and $M$ denote the pruning and meta learning techniques.
}
\end{table}
 %  ########################################
\subsection{Ablation Study}
In order to verify the effectiveness of the pruning and meta learning modules, we conduct ablation evaluation on the proposed MetaSD on the FB15k-237 dataset. As shown in Table~\ref{ablation}, we can observe that the meta learning technique has great impact on the performance of the proposed MetaSD. This is because that meta learning can make the source network transfer rich knowledge to the pruned network effectively in the self-distillation process. In addition, the improvement of the self-pruning strategy is also significant since self-pruning can help the model learn discriminative representations and deal with the long-tail samples.  It is no surprise that combining both factors achieves the best performance on in terms of four evaluation metrics.

%the results are not as good as our MetaSD due to its insensitivity to long-tail samples. Without meta learning in self-distillation, the results are slightly worse because of the decline in the knowledge transfer ability of the source model. If we discard both modules, which backs to the deep mutual learning(DML), we can acquire 0.373 in MRR, suggesting that MetaSD can greatly compress the model without losing performance compared to DML.

\subsection{Generalization}
To demonstrate the robustness of our framework, we also implement MetaSD on two additional backbone models (e.g., CP and RESCAL).
These two backbone models are implemented and initialized by following the paper~\citep{zhang2020duality}.
As shown in Table~\ref{generation}, our compressed model achieves substantially better performance than the larger baseline models based on two different backbone models.

\begin{table}
\centering
\resizebox{0.48\textwidth}{!}{
\begin{tabular}{p{3.2cm}P{1.2cm}P{1.2cm}P{1.2cm}P{1.2cm}P{1.2cm}}
    \toprule
    % \hline
    \textbf{Model} & \textbf{MRR} & \textbf{H@1} & \textbf{H@3} & \textbf{H@10}&\textbf{Size} \\
    \hline\hline
    CP &0.333&0.247&0.360&0.508& 50M\\
    RESCAL & 0.353 & 0.264 & 0.383 &0.528 & 125M\\
    \rowcolor{gray!20}\textbf{MetaSD-CP} &0.367 & 0.270&0.396&0.557& 5M\\
    \rowcolor{gray!20}\textbf{MetaSD-RESCAL} &0.372&0.276&0.405&0.561 & 12.5M\\
    
    % \hline
    % \rowcolor{gray!20}MetaSD-$\mathcal{T}$ & 0.468 & 0.380 & 0.510 & 0.642\\
    % \rowcolor{gray!20}MetaSD & \textbf{0.471} & \textbf{0.381} & \textbf{0.512} & \textbf{0.646} \\
    \bottomrule
\end{tabular}
}
\caption{\label{generation}
Results of MetaSD on FB15k-237 by using different backbone models.
}
\end{table}
% #######################################
\section{Conclusion}
In this paper, we proposed a self-distillation framework with meta learning for graph embedding compression. Concretely, we proposed a one-step meta self-distillation method for distilling comprehensive knowledge from the source model to the pruned model, where the two models co-evolved in a dynamic manner during training. Experimental results showed that our model achieved competitive performance compared to strong baseline methods, while being 10x smaller than baseline methods. 

\section*{Acknowledgements}
This work was partially supported by National Key R\&D Program of China (No. 2019YFB2102500), National Natural Science Foundation of China (No. 61906185), Youth Innovation Promotion Association of CAS China (No. 2020357), Shenzhen Science and Technology Innovation Program (Grant No. KQTD20190929172835662), Shenzhen Basic Research Foundation (No. JCYJ20210324115614039 and No. JCYJ20200109113441941). 
%This work was supported by Alibaba Group through Alibaba Innovative Research Program.

\section*{Limitations}
To better understand the limitations of the proposed model, we carry out an analysis of the error predictions made by MetaSD. In particular, we primarily analyze the relations in the FB15k-237 test set, whose MRR scores are less than 0.2. Most of the incorrectly predicted relations are the ``location'' and ``relationships'' related relation types, such as place of birth/death, spouse, and sibling. We reveal several reasons of the bad cases, which can be divided into two primarily categories. First, MetaSD fails to predict some instances that require the multi-hop reasoning to get the correct answers, since our model does not consider the complex multi-hop paths during the knowledge graph representation learning. Second, MetaSD fails to predict some instances, where there are a large number of candidate entities to reason for a relation type (e.g., the location relation). One possible solution is to devise a two-step ranking method by filtering most of the irrelevant entities in a coarse-grained way and then distinguish the confusing entities with a fine-grained method. 

\bibliography{anthology}%,custom}
\bibliographystyle{acl_natbib}

% \appendix

% \section{Example Appendix}
% \label{sec:appendix}

% This is a section in the appendix.

\end{document}